\documentclass[10pt,twocolumn,letterpaper]{article}
\usepackage{authblk}
\usepackage[pagenumbers]{iccv} 

\definecolor{iccvblue}{rgb}{0.21,0.49,0.74}

\usepackage[pagebackref,breaklinks,colorlinks,allcolors=iccvblue]{hyperref}
\usepackage{multirow}
\usepackage{stmaryrd}
\def\paperID{*****} 
\def\confName{ICCV}
\def\confYear{2025}

\title{DGOcc: Depth-aware Global Query-based Network\\for Monocular 3D Occupancy Prediction}

\author[]{Xu Zhao}    

\author{Pengju Zhang}

\author{Bo Liu}

\author[]{Yihong Wu\thanks{Corresponding Author. Email: \texttt{yhwu@nlpr.ia.ac.cn}}}

\affil[]{State Key Laboratory of Multimodal Artificial Intelligence, Institute of Automation, \\Chinese Academy of Sciences, Beijing, China}     

\begin{document}
\maketitle
\begin{abstract}

Monocular 3D occupancy prediction, aiming to predict the occupancy and semantics within interesting regions of 3D scenes from only 2D images, has garnered increasing attention recently for its vital role in 3D scene understanding. Predicting the 3D occupancy of large-scale outdoor scenes from 2D images is ill-posed and resource-intensive. In this paper, we present \textbf{DGOcc}, a \textbf{D}epth-aware \textbf{G}lobal query-based network for monocular 3D \textbf{Occ}upancy prediction. We first explore prior depth maps to extract depth context features that provide explicit geometric information for the occupancy network. Then, in order to fully exploit the depth context features, we propose a Global Query-based (GQ) Module. The cooperation of attention mechanisms and scale-aware operations facilitates the feature interaction between images and 3D voxels. Moreover, a Hierarchical Supervision Strategy (HSS) is designed to avoid upsampling the high-dimension 3D voxel features to full resolution, which mitigates GPU memory utilization and time cost. Extensive experiments on SemanticKITTI and SSCBench-KITTI-360 datasets demonstrate that the proposed method achieves the best performance on monocular semantic occupancy prediction while reducing GPU and time overhead.
\end{abstract}    
\section{Introduction}
\label{sec:intro}

3D occupancy prediction has gained widespread attention in the past few years. It aims to construct the holistic scene by predicting occupancy and semantics for each voxel of the 3D volumetric grid within the 3D scene. Predicting the 3D occupancy of outdoor scenes is significant for autonomous vehicles to holistically understand 3D scenes and avoid potential obstacles \cite{occnet}. The occupancy representation provides finer-grained details and superior zero-shot capability than traditional representations \cite{pop3d, ovo, zhang2024survey} like 3D Bounding Box and Bird’s-Eye-View (BEV). Despite its great promise for 3D scene understanding, 3D occupancy prediction faces substantial obstacles, such as incomplete observations and high demands for GPU resources. These problems make it particularly challenging to reconstruct complete scenes from partial measurements of sensors, especially when GPU memory is limited.

\begin{figure}
    \centering
    \begin{subfigure}{0.39\linewidth}
        \includegraphics[width=\linewidth]{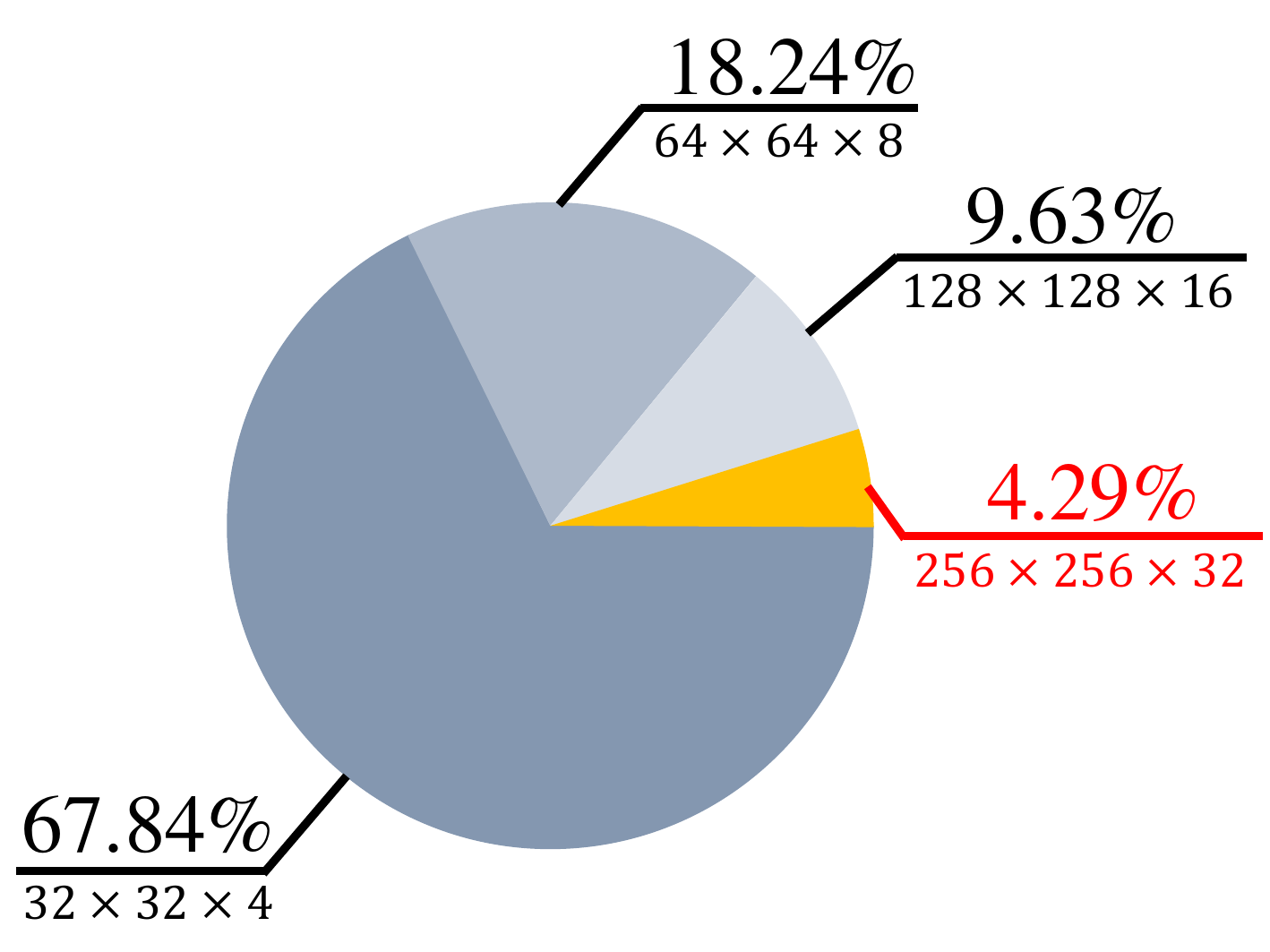}
        \caption{Sector of voxels}
        \label{fig:sector}
    \end{subfigure}
    \begin{subfigure}{0.6\linewidth}
        \includegraphics[width=\linewidth]{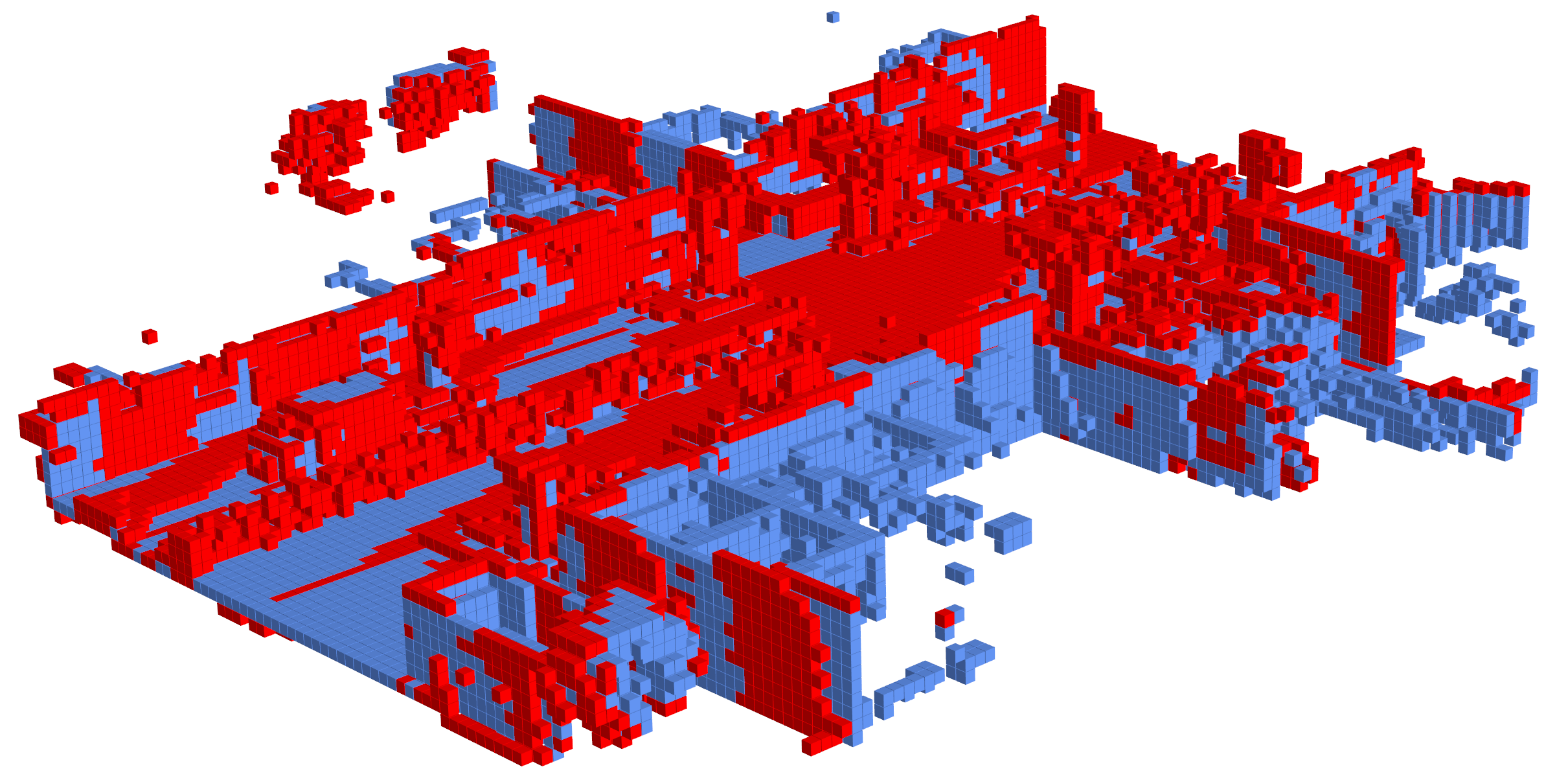}
        \caption{Voxels for subdivision}
        \label{fig:occviz}
    \end{subfigure}
    \caption{Statistical results of the 3D semantic voxel ground truth in SemanticKITTI validation set. (a) The chart shows the percentage of voxels that don't require further subdivision at different resolutions. (b) The red voxels should be subdivided at \(128\times 128\times 16\) resolution while the blue ones don't need.}
    \label{fig:intro}
\end{figure}

Under the continuous efforts of research scholars, considerable excellent semantic occupancy prediction methods have emerged. Earlier solutions \cite{lmscnet,s3cnet} depend on LiDAR points as inputs, but LiDAR points are expensive to acquire and deficient in providing textural information. Given the cheaper availability and richer visual cues of cameras, recent research trends have turned toward monocular solutions. MonoScene \cite{monoscene} is the first to recover the 3D scene occupancy solely from 2D perspective images. VoxFormer \cite{voxformer} adopts a pre-trained depth estimation model to provide depth prior. Symphonies \cite{symphonies} explores sparse instance queries to facilitate the feature interaction between 2D images and 3D volumetric grid. However, these methods \cite{depthssc, h2gfomer, symphonies} only use the estimated depth maps to provide position prior for queries but neglect rich depth context cues abundant in depth maps. This explicit depth context information can enhance feature diversity and help resolve 2D-to-3D ambiguity. Besides, large-scale outdoor scenes, which consist of complicated components with distinct shapes and sizes, are difficult to model with limited receptive field and scale-agnostic design. This also should be considered when predicting 3D occupancy.

Another troublesome issue is the high GPU and time requirements at training and inference stages. To solve this problem, TPVFormer \cite{tpvformer} proposes a memory-friendly tri-perspective view representation. FlashOcc \cite{flashocc} and FastOcc \cite{fastocc} adopt efficient 2D computational operations on BEV representation. These methods usually need to upsample the high-dimension 3D voxel features to the same resolution as the 3D semantic voxel ground truth before predicting semantics. We observe that the resolution of 3D semantic voxel ground truth is extremely high in related datasets. For example, it's up to \(256 \times 256\times 32\) in SemanticKITTI \cite{semantickitti}. Therefore, the upsampling process brings considerable memory usage and computational overhead. Statistics of the 3D semantic voxel ground truth in SemanticKITTI dataset are shown in Figure~\ref{fig:intro}. The statistics focus on voxels that have no demand for further subdivision at \(128 \times 128\times 16\) resolution, which means the semantic labels are consistent within a voxel. As can be seen from Figure~\ref{fig:sector}, only 4.29\% of the 3D voxels at \(128 \times 128\times 16\) resolution require upsampling to \(256 \times 256\times 32\) resolution. In Figure~\ref{fig:occviz}, the voxels that should be subdivided at \(128\times128\times16\) resolution are indicated in red. These voxels are usually located at the semantics border where semantics vary drastically.

In this paper, we present DGOcc, a depth-aware global query-based network for monocular 3D occupancy prediction. Firstly, 2D image features are augmented by injecting explicit depth context information that is extracted from prior depth maps. The explicit depth context information makes the 2D features depth-perceptive and diverse. Then we propose a Global Query-based Module, aiming to fully exploit the 2D depth-aware features. The collaboration of attention mechanisms and scale-aware operations not only expands queries' receptive field to the whole 3D scene but also enables multi-scale modeling, thus facilitating the interaction of depth-aware features. Finally, we introduce a Hierarchical Supervision Strategy to circumvent the resource-intensive operation of upsampling all 3D voxel features to high resolution. High-level voxel features are obtained by splitting carefully selected low-level voxel features. The voxel features at different levels are supervised by their tailored losses.

Exhaustive experiments have been conducted on challenging SemanticKITTI \cite{semantickitti} and SSCBench-KITTI-360 \cite{kitti360, SSCBench} datasets to verify our method. The proposed method achieves a satisfactory performance of 16.14 and 19.46 mIoU respectively while reducing GPU memory and time consumption.

Our main contributions within this work are summarized as follows:
\begin{itemize}
    \item Prior depth maps are explored as a source of features, endowing explicit depth context information for the network.
    \item A Global Query-based Module is proposed, which captures long-range dependence and boosts the interaction of 2D depth-aware features.
    \item A Hierarchical Supervision Strategy is introduced to remarkably decrease GPU memory utilization and time cost at both training and inference stages.
    \item Based on the aforementioned approaches, comprehensive experiments are conducted on SemanticKITTI and SSCBench-KITTI-360 benchmarks, confirming that the proposed method achieves better performance while having less GPU and time overhead.
\end{itemize}

\section{Related work}
\label{sec:related}

\subsection{Vision-based BEV Perception}

Given the convenience of deployment, cost-efficiency, and expression capability of BEV representation, it is widely applied in vision-based 3D perception tasks. Many researchers propose methods to construct BEV features from perspective image features. LSS \cite{lss} produces 3D pseudo point cloud features by performing outer product between the 2D context features and the discrete depth distribution generated from 2D image features. The 3D pseudo point cloud features are then voxelized into BEV features. BEVDET \cite{bevdet} applies LSS to the 3D object detection task. BEVDepth \cite{bevdepth} further introduces LiDAR points to explicitly supervise the depth distribution predictions of LSS. BEVFormer \cite{bevformer} projects the BEV features into the perspective image features and implements 2D deformable attention to update the BEV features. In this paper, we replace BEV with occupancy representation for its stronger scene modeling capability but continue to adopt techniques used in BEV, such as Deformable Attention \cite{deformable}.

\begin{figure*}[t]
    \centering
    \includegraphics[width=\linewidth]{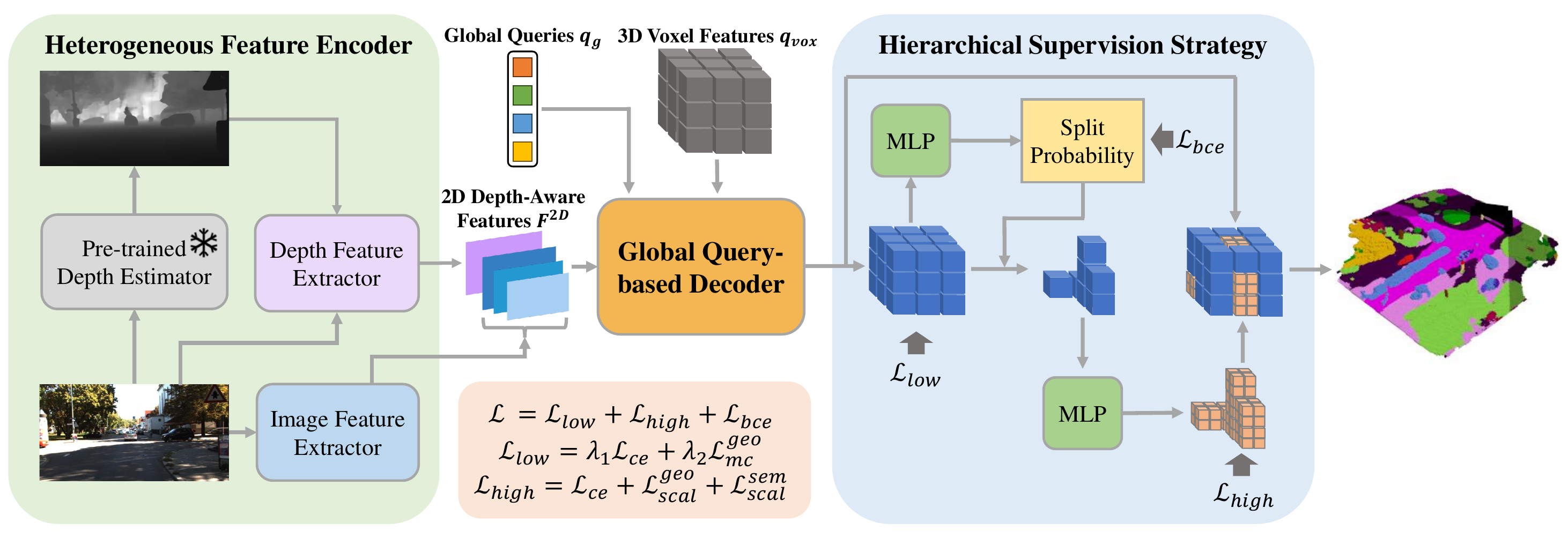}
    
    \caption{Overview of DGOcc. Given input images, Pre-trained Depth Estimator first estimates a depth map for each image. Then, Heterogeneous Feature Encoder is employed to extract multi-scale image features and single-scale depth context features. The two features constitute the 2D depth-aware features. Global Query-based Decoder propagates information from 2D depth-aware features to 3D voxel features with global queries. The resulting 3D voxel features are finally fed to the Hierarchical Supervision Strategy module to generate hierarchical occupancy predictions.}
    \label{fig:overview}

\end{figure*}

\subsection{3D Occupancy Prediction}

3D occupancy prediction aims to understand complex 3D scenes by predicting semantics for each voxel of 3D volumetric grid, which was first proposed by SSCNet \cite{sscnet}. Earlier research \cite{ssasc,scpnet,pointocc} employs LiDAR data as input for occupancy prediction. Considering the expense of LiDAR sensors, some studies \cite{surroundocc,ndcscene} explore the possibility of utilizing camera data. MonoScene \cite{monoscene} is the first 3D occupancy prediction work to take exclusively RGB images as inputs, which proposes FloSP to construct 3D volumetric features from multi-scale image features and 3D CRP to enhance spatial-semantic awareness. OccFormer \cite{occformer} proposes a dual-path transformer encoder to generate multi-scale 3D volumetric features and employs Mask2Former-like \cite{mask2former} occupancy decoder to predict semantics. VoxFormer \cite{voxformer} introduces depth prior by using an off-the-shelf depth estimation model to predict a depth map for each image and follows an MAE-like \cite{mae} procedure to diffuse the features from visible query proposals to the whole scene. MonoOcc \cite{monoocc} follows up VoxFormer, introducing 2D semantic segmentation subtask and distillation framework to enhance performance. HASSC \cite{hassc} proposes a hardness-aware scheme to focus on the voxels that are hard to distinguish, and introduces a self-distillation training strategy to improve semantic occupancy prediction. Symphonies \cite{symphonies} facilitates feature interactions and captures global scene context by introducing sparse instance queries. These methods don't fully exploit the depth context information available in prior depth maps. In this paper, we treat prior depth maps as a source of features and equip the network with long-range dependence and scale-aware capabilities.

\subsection{Lightweight Occupancy}

Given 3D occupancy prediction suffers from large GPU memory consumption and is not real-time, many studies focus on lightweight occupancy research. FlashOcc \cite{flashocc} and FastOcc \cite{fastocc} adopt 2D BEV representation for occupancy prediction, resulting in height information loss. TPVFormer \cite{tpvformer} proposes the tri-perspective view representation, striking a balance between BEV and 3D occupancy representation. CTF-Occ \cite{occ3d} employs a coarse-to-fine pipeline and only selects the predicted occupied voxels to interact with image features. SparseOcc \cite{sparseocc} also uses a coarse-to-fine pipeline but directly discards the predicted empty voxels. OccFusion \cite{occfusion} adopts an active coarse-to-fine pipeline to choose voxels with large entropy for further feature interaction. Although selecting occupied voxels significantly reduces computational burden, it can still be further optimized by selecting voxels that should be subdivided whose quantity is smaller. In this paper, we concentrate on selecting voxels that should be subdivided and propose a multi-class version of losses to preserve as much information as possible in these voxels, ensuring that these voxels no longer require feature interaction after subdividing.

\section{Methodology}

The overall architecture of DGOcc is illustrated in Figure~\ref{fig:overview}. Our DGOcc is comprised of three components: Heterogeneous Feature Encoder, Global Query-based Decoder, and Hierarchical Supervision Strategy.

In Heterogeneous Feature Encoder, we first employ a pre-trained Depth Estimator to generate depth maps from input images. The input images and estimated depth maps are jointly fed into two different Feature Extractors to extract 2D depth-aware features \(F^{2D}\). The 2D depth-aware features include both multi-scale image features \(F^{image}\) and single-scale depth context features \(F^{depth}\).

\begin{figure*}[t]
    \centering
    \includegraphics[width=0.9\linewidth]{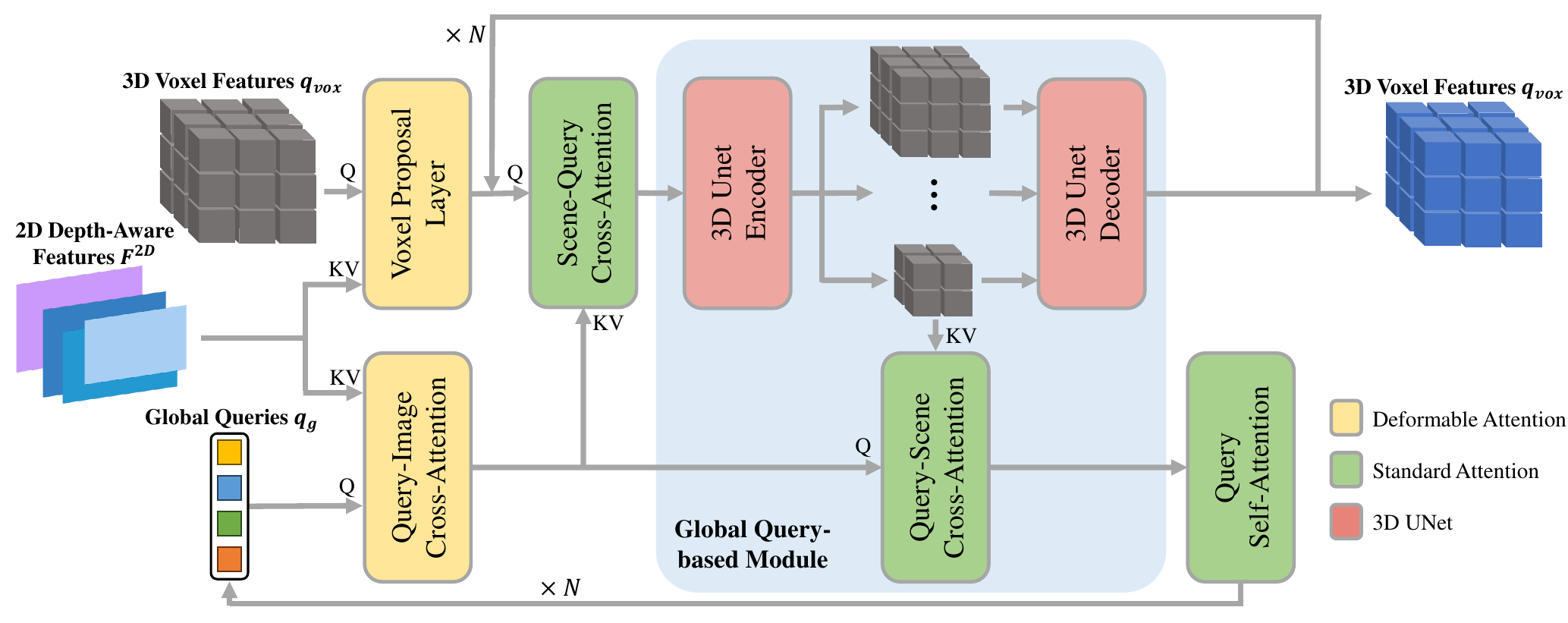}
    \caption{Illustration of the Global Query-based Decoder. 3D voxel features and global queries are first initialized with 2D depth-aware features in respective modules. Then a multi-scale and global aware paradigm exchanges information between 3D voxel features and global queries. After \(N\) iterations, 3D voxel features saturated with geometric and semantic cues are used for occupancy prediction.}
    \label{fig:network}
\end{figure*}

In Global Query-based Decoder, global queries \(q_g \in \mathbb{R}^{N\times C}\) and 3D voxel features \(q_{vox}\in \mathbb{R}^{C\times X\times Y\times Z} \) are both learnable embeddings, where \(C\) refers to the embedding channels, \(N\) denotes the number of global queries, and \((X,Y,Z)\) is the shape of scene grid. The 3D voxel features are initialized with prior depth points by aggregating 2D depth-aware features. We iteratively adopt a series of feature interaction operations to propagate 2D depth-aware features to 3D voxel features by global queries, generating low-level 3D voxel features \(q_{low}\in \mathbb{R}^{C\times X\times Y\times Z} \) with rich geometric and semantic information.

Finally, our Hierarchical Supervision Strategy is designed to choose low-level 3D voxel features that require subdivision to split into full resolution, forming high-level 3D voxel features \(q_{high} \in \mathbb{R}^{C\times 8K} \), where \(K\) indicates the number of selected low-level 3D voxel features. The resulting high-level 3D voxel features and low-level 3D voxel features are supervised by respective tailored losses, which avoids upsampling all voxels to high resolution thus extremely reducing the GPU memory utilization and time cost. We detail each component in subsequent sections.

\subsection{Heterogeneous Feature Encoder}

To align with previous methods \cite{voxformer,symphonies}, we employ the pre-trained Mobilestereonet \cite{mobilestereonet} as depth estimator to provide pixel-wise depth \(D\in \mathbb{R}^{H\times W} \) for input images. A pre-trained MaskDINO \cite{maskdino} encoder is adopted as Image Feature Extractor to extract multi-scale image features \(F^{image}=\{F^{image}_i \in \mathbb{R}^{C\times H_i \times W_i} | i=1,2,..,s \}\) from input RGB images, where \(s\) is the scale number of image features, \(C\) denotes the channel number and  \((H_i,W_i)\) represents the resolution of \(i\)-th scale image feature map.

Inspired by KYN \cite{kyn}, enriching feature diversity with other information can boost the network's modeling capability. Monocular image features comprise both textural and semantic information, but lack explicit depth context information, which is critical to 3D scene recovery from 2D images. To provide depth context information, we propose a Depth Feature Extractor that takes images and estimated depth maps as inputs. Specifically, the Depth Feature Extractor consists of a lightweight ResNet18-like \cite{resnet} backbone to first extract multi-scale depth context features and a SECONDFPN \cite{second} neck to fuse them, then followed by a \(1 \times 1\) convolution layer to align channel dimension with the image features. Finally, the single-scale depth context features \(F^{depth} \in \mathbb{R}^{C\times H_d \times W_d}\) are simply appended to multi-scale image features \(F^{image}\), forming the 2D depth-aware features \( F^{2D}\). Although the fusion method is straightforward, it facilitates subsequent feature interactions.

\begin{table*}[t]
    \centering
    \small
    \setlength{\tabcolsep}{0.5mm}
    \begin{tabular}{l|cc|ccccccccccccccccccc} \toprule 
         Method & IoU  & mIoU & \rotatebox[origin=l]{90}{road} & \rotatebox[origin=l]{90}{sidewalk} & \rotatebox[origin=l]{90}{parking} & \rotatebox[origin=l]{90}{other-grnd.} & \rotatebox[origin=l]{90}{building} & \rotatebox[origin=l]{90}{car} & \rotatebox[origin=l]{90}{truck} & \rotatebox[origin=l]{90}{bicycle} & \rotatebox[origin=l]{90}{motorcycle} & \rotatebox[origin=l]{90}{other-veh.} & \rotatebox[origin=l]{90}{vegetation} & \rotatebox[origin=l]{90}{trunk} & \rotatebox[origin=l]{90}{terrain} & \rotatebox[origin=l]{90}{person} & \rotatebox[origin=l]{90}{bicyclist} & \rotatebox[origin=l]{90}{motorcyclist} & \rotatebox[origin=l]{90}{fence} & \rotatebox[origin=l]{90}{pole} & \rotatebox[origin=l]{90}{traf.-sign} \\ \midrule 
         MonoScene\(^\ast\) & 34.16& 11.08& 54.70& 27.10& 24.80& 5.70 & 14.40& 18.80& 3.30& 0.50& 0.70& 4.40& 14.90& 2.40 & 19.50& 1.00& 1.40& 0.40& 11.10& 3.30& 2.10\\
         TPVFormer & 34.25& 11.26& 55.10& 27.20& 27.40& 6.50 & 14.80& 19.20& 3.70& 1.00& 0.50& 2.30& 13.90& 2.60 & 20.40& 1.10& 2.40& 0.30& 11.00& 2.90& 1.50\\
         VoxFormer\(^\dagger\) & \underline{43.21} & 13.41 & 54.10 & 26.90 & 25.10 & 7.30 & 23.50 & 21.70 & 3.60 & 1.90 & 1.60 & 4.10 & 24.40 & 8.10 & 24.20 & 1.60 & 1.10 & 0.00 & 13.10 & 6.60 & 5.70 \\
         OccFormer & 34.53& 12.32& 55.90& 30.30& \underline{31.50} & 6.50 & 15.70& 21.60& 1.20& 1.50& 1.70& 3.20& 16.80& 3.90 & 21.30& 2.20& 1.10& 0.20& 11.90& 3.80& 3.70\\
         MonoOcc & - & \underline{15.63} & \underline{59.10} & \underline{30.90} & 27.10 & 9.80 & 22.90 & \underline{23.90} & \textbf{7.20} & \textbf{4.50} & 2.40 & \textbf{7.70} & \underline{25.00} & 9.80 & 26.10 & \underline{2.80} & \textbf{4.70} & \underline{0.60} & \underline{16.90} & \underline{7.30} & \textbf{8.40} \\
         Symphonies& 42.19& 15.04& 58.40& 29.30& 26.90& \underline{11.70}& \textbf{24.70} & 23.60& 3.20& \underline{3.60} & \underline{2.60} & \underline{5.60} & 24.20& \underline{10.00} & 23.10& \textbf{3.20} & 1.90& \textbf{2.00} & 16.10& \textbf{7.70} & \underline{8.00} \\ 
         HASSC\(^\dagger\) & 42.87 & 14.38 & 55.30 & 29.60 & 25.90 & 11.30 & 23.10 & 23.00 & 2.90 & 1.90 & 1.50 & 4.90 & 24.80 & 9.80 & \underline{26.50} & 1.40 & \underline{3.00} & 0.00 & 14.30 & 7.00 & 7.10 \\ \midrule
         Ours & \textbf{44.32} & \textbf{16.14} & \textbf{63.70} & \textbf{33.00} & \textbf{32.60} & \textbf{12.70} & \underline{23.70} & \textbf{24.00} & \underline{6.40} & 3.50 & \textbf{2.60} & 5.00 & \textbf{25.40} & \textbf{10.10} & \textbf{27.60} & 2.20 & 1.80 & 0.50 & \textbf{17.60} & 7.20 & 7.00 \\ \bottomrule
    \end{tabular}
    \caption{Quantitative results on SemanticKITTI hidden test set. \(^\ast\) denotes the reproduced results in related papers \cite{tpvformer, occformer}. \(^\dagger\) represents the results with temporal inputs. The best and the second-best results among all methods are marked bold and underlined respectively.}
    \label{tab:semantictest}
\end{table*}

\subsection{Global Query-based Decoder}

Our Global Query-based Decoder extends instance queries \cite{symphonies} to global queries by expanding the receptive field of queries from the local region to the whole scene. Additionally, scale-aware structures are added to the network, enabling multi-scale modeling. The entire pipeline of Global Query-based Decoder is illustrated in Figure~\ref{fig:network}. Voxel Proposal Layer first uses the prior depth maps to select voxel proposals. Then Deformable Attention \cite{deformable} is employed to aggregate the 2D depth-aware features \(F^{2D}\) for the features of voxel proposals. In Query-Image Cross-Attention, the same deformable attention operation propagates information from 2D depth-aware features \(F^{2D}\) to global queries \(q_g\). Then Scene-Query Cross-Attention enables global feature aggregation from global queries to the 3D voxel features within the input images' field of view. Subsequently, 3D voxel features \(q_{vox}\) and global queries \(q_g\) are together input into the Global Query-based Module. The designed module expands the receptive field and enables multi-scale modeling during feature interactions. The output global queries exchange global context information in Query Self-Attention. After \(N\) iterations, 3D voxel features are saturated with rich geometric and semantic information. In the following part, we describe our Global Query-based Module in detail.

Large-scale outdoor scenes are comprised of diverse foreground instances and backgrounds, ranging from small traffic signs to large buildings. Therefore, it’s difficult to represent holistic scenes by treating all scene constituents on the same scale. We employ a lightweight 3D UNet to capture multi-scale spatial relations within 3D voxel features, which contributes to the hierarchical understanding of scenes. Additionally, it also propagates features from voxel proposals to all voxels. This is crucial to hallucinate geometric and semantic details beyond the input images’ visible region. Specifically, 3D voxel features are first input into 3D UNet Encoder to generate multi-scale voxel features \(q_{ms}=\{q_{vox}^i\in \mathbb{R}^{C\times X_i\times Y_i\times Z_i} | i=1,..,S\}\), where \(S\) is the index of the lowest scale. This process are formulated as \( q_{ms}=\text{UNetEncoder}(q_{vox})\). Then the multi-scale voxel features are upsampled and concatenated with the same scale features in 3D UNet Decoder to construct the updated 3D voxel features: \(q_{vox}=\text{UNetDecoder}(q_{ms})\). 

Queries serve as the global medium to facilitate the feature interaction between images and 3D voxels. To expand the receptive field of queries from the local region to the entire scene, we adopt cross-attention to integrate scene information for queries from voxel features in Query-Scene Cross-Attention. However, the high resolution of 3D voxel features leads to excessive GPU memory and computation overhead. To address this issue, cross-attention is only employed at the lowest resolution of multi-scale voxel features, expressed as \(q_{g} = \text{CrossAttn}(q_{g}, q_{vox}^{S})\).

\begin{table*}[t]
    \centering
    \small
    \setlength{\tabcolsep}{0.5mm}
    \begin{tabular}{l|cc|cccccccccccccccccc} \toprule 
         Method & IoU  & mIoU & \rotatebox[origin=l]{90}{car} & \rotatebox[origin=l]{90}{bicycle} & \rotatebox[origin=l]{90}{motorcycle} & \rotatebox[origin=l]{90}{truck} & \rotatebox[origin=l]{90}{other-veh.} & \rotatebox[origin=l]{90}{person} & \rotatebox[origin=l]{90}{road} & \rotatebox[origin=l]{90}{parking} & \rotatebox[origin=l]{90}{sidewalk} & \rotatebox[origin=l]{90}{other-grnd.} & \rotatebox[origin=l]{90}{building} & \rotatebox[origin=l]{90}{fence} & \rotatebox[origin=l]{90}{vegetation} & \rotatebox[origin=l]{90}{terrain} & \rotatebox[origin=l]{90}{pole} & \rotatebox[origin=l]{90}{traf.-sign} & \rotatebox[origin=l]{90}{other-struct} & \rotatebox[origin=l]{90}{other-obj.}  \\ \midrule 
         MonoScene & 37.87 & 12.31 & 19.34 & 0.43 & 0.58 & 8.02  & 2.03 & 0.86 & 48.35 & 11.38 & 28.13 & 3.32 & 32.89 & 3.53 & 26.15 & 16.75 & 6.92 & 5.67 & 4.20 & 3.09  \\
	TPVFormer & 40.22 & 13.64 & 21.56 & 1.09 & 1.37 & 8.06  & 2.57 & 2.38 & 52.99 & 11.99 & 31.07 & 3.78 & 34.83 & 4.80 & 30.08 & 7.52 & 7.46 & 5.86 & 5.48 & 2.70  \\
	OccFormer & 40.27 & 13.81 & 22.58 & 0.66 & 0.26 & 9.89  & 3.82 & 2.77 & 54.30       
			& 13.44 & 31.53 & 3.55 & \underline{36.42} & 4.80 & 31.00 & \underline{19.51} & 7.77 & 8.51 & 6.95 & 4.60  \\
	VoxFormer & 38.76 & 11.91 & 17.84 & 1.16 & 0.89 & 4.56  & 2.06 & 1.63 & 47.01
			& 9.67  & 27.21 & 2.89 & 31.18 & 4.97 & 28.99 & 14.69 & 6.51 & 6.92 & 3.79 & 2.43  \\
	Symphonies & \underline{44.12} & \underline{18.58} & \textbf{30.02} & \underline{1.85} & \textbf{5.90} & \textbf{25.07} & \textbf{12.06} & \textbf{8.20} & \underline{54.94} & \underline{13.83} & \underline{32.76} & \textbf{6.93} & 35.11 & \textbf{8.58} & \textbf{38.33} & 11.52 & \textbf{14.01} & \underline{9.57} & \textbf{14.44} & \textbf{11.28} \\ \hline
         Ours & \textbf{46.10} & \textbf{19.46} & \underline{28.32} & \textbf{4.24} & \underline{4.92} & \underline{15.85} & \underline{9.45} & \underline{7.43} & \textbf{62.23} & \textbf{18.41} & \textbf{40.20} & \underline{5.37} & \textbf{40.36} & \underline{8.37} & \underline{35.93} & \textbf{23.68} & \underline{13.61} & \textbf{16.14} & \underline{9.42} & \underline{6.35} \\ \bottomrule
    \end{tabular}
    \caption{Quantitative results on SSCBench-KITTI-360 test set. The results for counterparts are provided in SSCBench \cite{SSCBench}. The best results are in bold and the second-best results are underlined.}
    \label{tab:kitti360}
\end{table*}

\subsection{Hierarchical Supervision Strategy}
Training occupancy networks is usually a tedious process and places a tremendously high demand for GPU memory. Previous methods need to upsample all of the 3D voxel features to high resolution before predicting semantics, leading to tremendous GPU consumption. To solve this problem, we propose a Hierarchical Supervision Strategy, as depicted in Figure~\ref{fig:overview}. Specifically, we directly use a low-level semantic head to predict semantic logits for 3D voxels at low resolution, then employ a low-level loss \(\mathcal{L}_{low}\) on low-resolution predictions to constrain the semantic predictions. 
Given the fact that a low-level voxel may be composed of many high-level voxels with different classes, directly using one-hot labels interpolated by ground truth would cause information loss. Instead, we first count the number of each class within a low-level voxel, then normalize the number to serve as low-level label. Because our low-level labels are multi-class labels, we extend Scene-Class Affinity Loss \cite{monoscene} to multi-class representation. The multi-class label and predicted probability of class c for voxel \(i\) are denoted as \(p_{i,c}\) and  \(\hat{p}_{i,c}\) respectively, then our multi-class version of Scene-Class Affinity Loss is expressed as follows:
\begin{equation}
    \begin{aligned}
        \mathcal{L}_{mc}(\hat{p},p) & =-\frac{1}{C}\sum_{c=1}^C(P_c(\hat{p},p)+R_c(\hat{p},p)+S_c(\hat{p},p)), \\
        P_c(\hat{p},p) & =\log \frac{\sum_i \hat{p}_{i,c} \llbracket {p_{i,c} > 0} \rrbracket }{\sum_i \hat{p}_{i,c}}, \\
        R_c(\hat{p},p) & = - \left| \log \frac{\sum_i \hat{p}_{i,c} \llbracket {p_{i,c} > 0} \rrbracket }{\sum_i p_{i,c}} \right|, \\
        S_c(\hat{p},p) & =\log \frac{ \sum_i (1-\hat{p}_{i,c}) (1-\llbracket {p_{i,c} > 0} \rrbracket) }{\sum_i (1-\llbracket {p_{i,c} > 0} \rrbracket)}.
    \end{aligned}
\end{equation}
We utilize a combination of geometry \(\mathcal{L}_{mc}^{geo}\) and weighted cross-entropy loss as our low-level loss:
\begin{align}
    \mathcal{L}_{low}=\lambda_1 \mathcal{L}_{ce}+ \lambda_2 \mathcal{L}_{mc}^{geo}
    \label{eq:loss_low}
\end{align}
where \(\lambda_1\) and \(\lambda_2\) are the loss weights. 

In order to compensate for the loss of details, low-resolution 3D voxel features are input into MLP to predict per-voxel split probability. The split probability is supervised by a binary cross-entropy loss \(\mathcal{L}_{bce}\), where the ground truth label is 1 if the voxel requires subdivision and 0 otherwise. Then top \(K\) features with the highest probabilities are selected and split to high resolution by another MLP. After a high-level semantic head, the high-level semantic logits are supervised by a high-level loss \(\mathcal{L}_{high}\). The high-level loss imposes more constraints on border voxels that are difficult to discriminate, thus refining the fine-grained details of scenes. We use weighted cross-entropy loss \(\mathcal{L}_{ce}\), Scene-Class Affinity loss \(\mathcal{L}_{scal}^{geo}\) and \(\mathcal{L}_{scal}^{sem}\) as our high-level loss. The high-level loss function is formulated as follows:
\begin{align}
    \mathcal{L}_{high}=\mathcal{L}_{ce}+\mathcal{L}_{scal}^{geo}+\mathcal{L}_{scal}^{sem}
\end{align}
The overall loss function is \(\mathcal{L} = \mathcal{L}_{high} + \mathcal{L}_{low}+\mathcal{L}_{bce}\). In the end, the low-level logits and high-level logits are combined as final semantic predictions.

\subsubsection{Discussion.}
Although many prior works have explored voxel sparsification to reduce computational burden, our proposed HSS still has its advantages. CTF-Occ \cite{occ3d} and SparseOcc \cite{sparseocc} select occupied voxels for feature interactions, while our method concentrates on voxels that should be subdivided. These voxels exist only at semantics border and have a smaller quantity than occupied voxels. Therefore, our method saves more computational resources. The multi-class version of losses employed on low-level voxels in our method incurs less information loss than corresponding one-hot version used in other methods \cite{surroundocc}. Compared with OccFusion \cite{occfusion} whose purpose is to select voxels with large entropy, our selected voxels are directly supervised after subdividing without further feature interaction. Because our designed multi-class version of losses can maintain sufficient information, i.e. only loses permutation information. We deem the lost permutation information can be recovered through high-level supervision without additional feature interactions. This design requires less computational operations, thus making it more efficient. Furthermore, our parameterized selection rule works better than its entropy-based selection rule as shown in Table~\ref{tab:lambda}, enabling our method to achieve satisfying performance without additional computational operations.
\section{Experiments}
\label{sec:experiments}

\subsection{Datasets and Metrics}

\noindent\textbf{Datasets: }
We carry out experiments to verify our method on SemanticKITTI \cite{semantickitti} and SSCBench-KITTI-360 \cite{kitti360, SSCBench} datasets. Both datasets provide semantic ground truth of interesting scene region in the form of 3D voxel grid with \(256\times 256\times 32\) resolution, whose voxel size is \(0.2m\), representing a range of \(51.2m\times 51.2m\times 6.4m\). The SemanticKITTI dataset consists of 22 urban driving scene sequences, with 9 allocated for training, 1 for validation, and 11 for testing. It provides RGB images with the resolution of \(1226\times 370\). The voxel grid is annotated with 20 classes (19 semantic classes and 1 free class). SSCBench-KITTI-360 comprises 9 sequences, among which 7 are used for training, 1 for validation, and 1 for testing, where RGB images have a resolution of \(1408\times 376\), encompassing 19 classes (18 semantic classes and 1 free class).

\noindent\textbf{Evaluation Metrics: }
Following previous methods \cite{monoscene, voxformer, symphonies}, we report the intersection over union (IoU) and the mean IoU (mIoU) for scene completion quality evaluation and semantic segmentation performance assessment, respectively.

\subsection{Implementation Details}
We train our model on 2 NVIDIA TITAN RTX GPUs for 30 epochs, with a batch size of 2 samples. We crop the input images to \(1220\times 370\) and \(1396\times 372\) for SemanticKITTI and SSCBench-KITTI-360 datasets respectively. Random horizontal flip augmentation is implemented for both datasets. The AdamW \cite{adam} optimizer is adopted with 2e-4 learning rate and 1e-4 weight decay. The WarmupMultiStepLR strategy is used with a factor of 0.01 at the beginning of 2 epochs to stabilize the training process. Learning rate reduction occurs by a factor of 0.1 at the 25th epoch. We set \(\lambda_1 = 1.0\), \(\lambda_2 = 0.3\) and \(K=15000\) for both training and inference stages. The Depth Feature Extractor is initialized by the weights from our Depth-Semantic Joint Pre-training. We reduce the learning rate for the Depth Feature Extractor by a factor of 0.2. For pre-training details, please refer to the appendix.

\begin{figure*}[t]
    \centering
    \includegraphics[width=\linewidth]{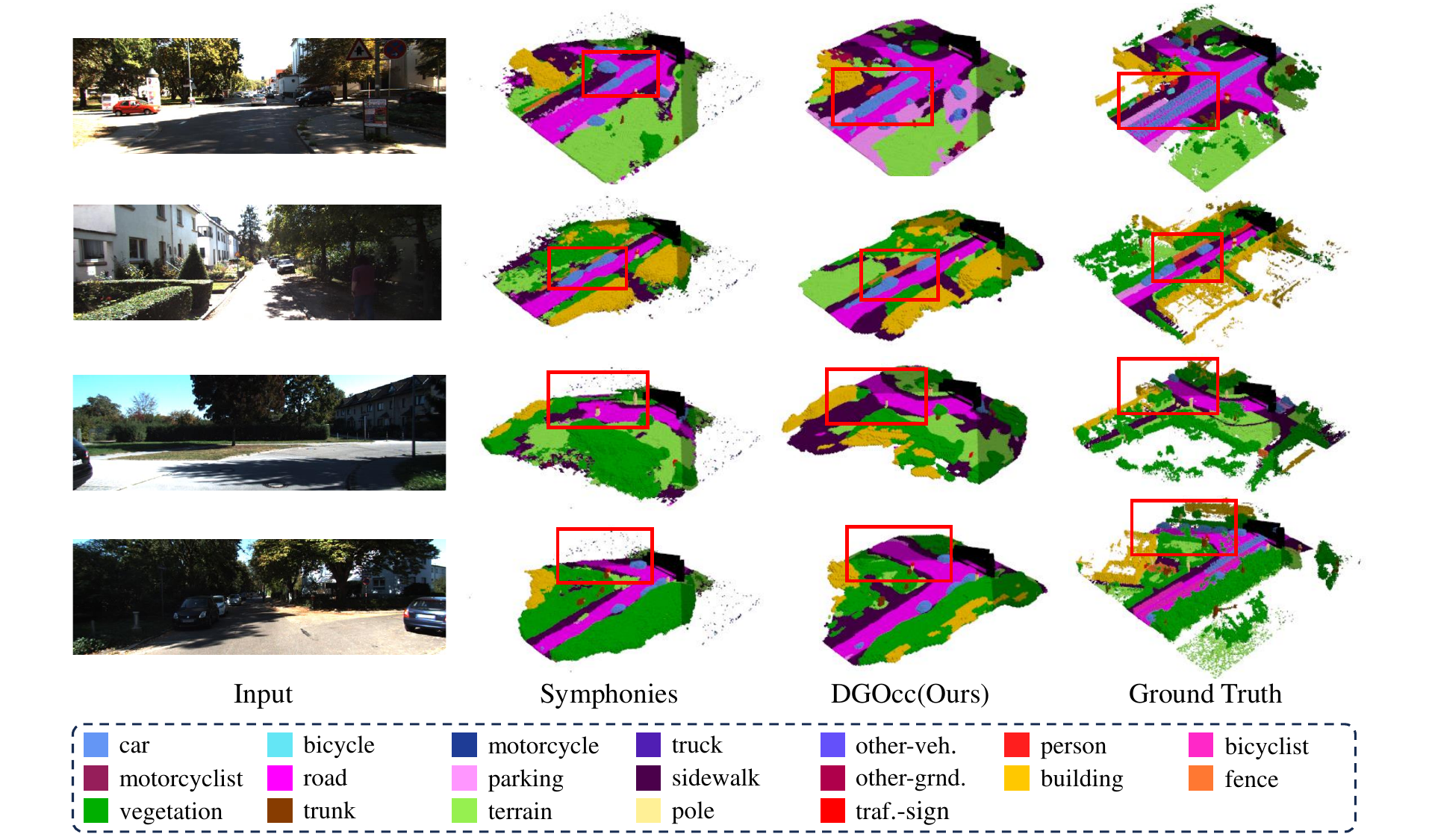}
    \caption{Qualitative results of Symphonies and DGOcc on SemanticKITTI validation set. DGOcc possesses enhanced capability in hallucinating unseen regions, thus recoverying more complete scene. Moreover, DGOcc is expert in distinguishing different objects' locations, for example cars.}
    \label{fig:visualization}
\end{figure*}

\begin{table}[hb]
    \centering
    \begin{tabular}{l|l|ccc} \toprule
    \multicolumn{2}{c|}{Method} & Symp. & Ours w/o HSS & Ours \\ \midrule
    \multirow{2}{*}{\rotatebox{90}{Train}} & Time (min)\(\downarrow\) & 58.93 & 49.01 & \textbf{28.00} \\ \cline{2-2}
    & Memory (G)\(\downarrow\) & 21.03 & 18.79 & \textbf{11.81} \\ \midrule
    \multirow{4}{*}{\rotatebox{90}{Inference}} & Time (ms)\(\downarrow\) & 251.83 & 272.05 & \textbf{238.85} \\ \cline{2-2}
    & Memory (G)\(\downarrow\) & 5.10 & 4.81 & \textbf{3.29}  \\ \cline{2-2}
    & IoU \(\uparrow\) & 41.92 & \textbf{45.13} & 44.98  \\ \cline{2-2}
    & mIoU\(\uparrow\) & 14.89 & \textbf{15.98} & 15.73  \\ \bottomrule
    \end{tabular}
    \caption{Comparison of efficiency on SemanticKITTI val set. For training, the time consumption for one epoch with 4 GPUs is evaluated. For inference, the time required to infer a single frame is recorded. HSS is our Hierarchical Supervision Strategy. Symp. represents the experiments of Symphonies that are conducted with official implements.}
    \label{tab:efficiency}
\end{table}

\subsection{Experimental Results}

\noindent\textbf{Quantitative Results: }
To demonstrate the effectiveness of our proposed method, we conduct extensive experiments on SemanticKITTI and SSCBench-KITTI-360 datasets. The results for 3D occupancy prediction on SemanticKITTI hidden test set and SSCBench-KITTI-360 test set are shown in Table~\ref{tab:semantictest} and Table~\ref{tab:kitti360}. Our method achieves 16.14 mIoU and 19.46 mIoU respectively, exhibiting the best performance. Our method surpasses Symphonies \cite{symphonies} on both IoU and mIoU metrics, and even possesses higher mIoU than methods that incorporate temporal information \cite{voxformer, hassc} or larger backbone \cite{monoocc}. As for per-class evaluation, the proposed method achieves the best or second-best performance on most of the classes, such as road, car, and fence.

Moreover, we evaluate the efficiency of our method, as presented in Table~\ref{tab:efficiency}. We compare our method with Symphonies \cite{symphonies} at both training and inference stages for its best performance among other methods. 
The results indicate that our approach extremely decreases the GPU memory and time cost at training stage, up to 43.84\% and 52.49\% respectively. 
We owe it to the avoidance of full-resolution feature upsampling. 
For the inference stage, our method also mitigates the GPU memory utilization and inference time cost. 
Apparently, the Hierarchical Supervision Strategy effectively minimizes the overhead with negligible performance degradation. The proposed method achieves a better performance than Symphonies, while demanding fewer GPU and time resources.

\begin{table}[ht]
    \centering
    \begin{tabular}{l|cc} \toprule 
         Method & IoU\(\uparrow\) & mIoU\(\uparrow\) \\ \midrule 
         Symphonies & 41.92  & 14.89 \\
         w/o GQ Decoder & 43.41 & 14.61 \\ 
         w/o Depth Feature Extractor & 44.55 & 15.65 \\ \midrule
         Ours & \textbf{44.98} & \textbf{15.73} \\ \bottomrule
    \end{tabular}
    \caption{Ablation study on architectural components. The experiments are conducted with the Hierarchical Supervision Strategy.}
    \label{tab:component}
\end{table}

\noindent\textbf{Qualitative Study: }
To intuitively demonstrate our method’s efficacy, qualitative results of the predicted semantic occupancy on the SemanticKITTI validation set are presented in Figure~\ref{fig:visualization}, where the resolution of the voxel predictions is \(256\times 256\times 32\). The red boxes in the first two rows show that our approach locates cars more accurately. The red boxes in the last two rows show that the proposed method has stronger hallucination capability and reconstructs more complete roads. The visualization results illustrate that our method achieves superior reconstruction of the scene structure and shows better performance in scene understanding.

\begin{table}[h]
    \centering
    \begin{tabular}{ccc|ccc} \toprule 
        \(K\) & \(\mathcal{L}_{mc}^{geo}\) & S.R. & IoU\(\uparrow\) & mIoU\(\uparrow\) & Rec.(\%) \\ \midrule 
        15000 &  & MLP & 42.34 & 15.15 & 41.04\\
        15000 & \checkmark & MLP & \textbf{44.55} & \textbf{15.65} & 41.46\\ 
        10000 & \checkmark & MLP & 44.49 & 15.23 & 32.82\\ 
        20000 & \checkmark & MLP & 44.45 & 15.29 & 48.16\\ 
        15000 & \checkmark & Entropy & 42.35 & 13.87 & 13.42\\ \bottomrule
    \end{tabular}
    \caption{Ablation results on the Hierarchical Supervision Strategy. The experiments are conducted without the Depth Feature Extractor. S.R. represents the Selection Rule while Rec. is the Recall of voxels that require subdivision.}
    \label{tab:lambda}
\end{table}

\subsection{Ablation Studies}
In this subsection, we carry out ablation analyses on SemanticKITTI validation set to verify the effectiveness of the individual components in our proposed method.

\noindent\textbf{Ablation on architectural components: }
As illustrated in Table~\ref{tab:component}, all components of our method contribute to the best performance. Compared with Symphonies \cite{symphonies}, our GQ Decoder places more emphasis on multi-scale cues and enlarges the receptive field, leading to a significant improvement on both IoU and mIoU. The Depth Feature Extractor introduces explicit depth context information that benefits 3D scene structure recovery from 2D images, thus the performance is enhanced further.

\noindent\textbf{Ablation on the Hierarchical Supervision Strategy: }
The ablation study for our proposed Hierarchical Supervision Strategy is presented in Table~\ref{tab:lambda}. 
The hyperparameter \(K\) controls the number of high-resolution voxels supervised by the high-level loss \(\mathcal{L}_{high}\). 
We observe that, the model performs best when \(K\) is around the voxel number that should be subdivided, e.g. 11246 (\(4.29\%\times 128\times 128\times 16 \)) as Figure~\ref{fig:sector} indicates. 
Although a larger \(K\) yields a higher recall of voxels that should be subdivided, it impairs the optimization of mIoU and leads to the increment of GPU memory. 
Our multi-class version of Scene-Class Affinity Loss \(\mathcal{L}_{mc}^{geo}\) notably improves the performance, especially IoU. 
This loss possesses stronger constraints on optimization than the weighted cross-entropy loss. 
As for the selection rule, "Entropy" is the selection rule used in OccFusion \cite{occfusion}, which calculates each voxel's entropy by its low-level semantic logits and treats the entropy as criteria to subdivide voxels. 
Our MLP method surpasses its non-parameter method on both IoU and mIoU because our method possesses explicit supervision from the ground truth of voxels that require subdivision. 
Furthermore, our method chooses more proper voxels that should be subdivided.
\section{Conclusion}

In this paper, we present DGOcc, a depth-aware global query-based monocular 3D occupancy prediction approach that achieves both effectiveness and efficiency. We first incorporate depth context features to benefit the 3D scene recovery from 2D images by leveraging estimated depth maps. Then we propose a Global Query-based Module to facilitate the interactions of depth-aware features between images and 3D voxels. Finally, to reduce GPU and time overhead, we introduce a Hierarchical Supervision Strategy that makes our method efficient while maintaining effectiveness. Exhaustive experiments demonstrate that our method attains the best performance on SemanticKITTI and SSCBench-KITTI-360 benchmarks and simultaneously reduces GPU and time consumption.

{
    \small
    \bibliographystyle{ieeenat_fullname}
    \bibliography{main}
}

\appendix

\section{Details of Pre-training}

\subsection{Depth-Semantic Joint Pre-training}
The prior depth maps are noisy and the Depth Feature Extractor is initialized randomly. Both issues impede the extraction of effective depth context features. To refine the imprecise depth maps and begin training the occupancy network from a favorable starting point, we implement a Depth-Semantic Joint Pre-training for Depth Feature Extractor. Specifically, after the single-scale depth context features \(F^{depth} \in \mathbb{R}^{C\times H_d \times W_d}\) are extracted, they are upsampled and used to predict per-pixel 2D depth distribution \(P \in \mathbb{R}^{dmax\times H \times W}\) and semantics \(S \in \mathbb{R}^{n\times H \times W}\), where \(dmax\) is the number of predefined discrete depth and \(n\) represents the class number. The per-pixel 2D depth is then calculated as \(D=\sum_{i=1}^{dmax} P_i d_i \in \mathbb{R}^{H \times W}\), where \(d_i\) is the \(i\)-th predefined discrete depth.

LiDAR points with semantic annotation are projected onto images to generate sparse 2D depth and semantic labels. We use smooth L1 loss \(\mathcal{L}_{smoothL1}\) and cross-entropy loss \(\mathcal{L}_{ce}\) to constraint the predicted depth and semantics respectively. We only implement the losses where the projected depth of LiDAR points exists. After pre-training, the resulting weights are used to initialize the Depth Feature Extractor for subsequent training.

\begin{table*}[htb]
    \centering
    \small
    \setlength{\tabcolsep}{0.5mm}
    \begin{tabular}{l|cc|ccccccccccccccccccc} \toprule 
         Method & IoU  & mIoU & \rotatebox[origin=l]{90}{road} & \rotatebox[origin=l]{90}{sidewalk} & \rotatebox[origin=l]{90}{parking} & \rotatebox[origin=l]{90}{other-grnd.} & \rotatebox[origin=l]{90}{building} & \rotatebox[origin=l]{90}{car} & \rotatebox[origin=l]{90}{truck} & \rotatebox[origin=l]{90}{bicycle} & \rotatebox[origin=l]{90}{motorcycle} & \rotatebox[origin=l]{90}{other-veh.} & \rotatebox[origin=l]{90}{vegetation} & \rotatebox[origin=l]{90}{trunk} & \rotatebox[origin=l]{90}{terrain} & \rotatebox[origin=l]{90}{person} & \rotatebox[origin=l]{90}{bicyclist} & \rotatebox[origin=l]{90}{motorcyclist} & \rotatebox[origin=l]{90}{fence} & \rotatebox[origin=l]{90}{pole} & \rotatebox[origin=l]{90}{traf.-sign} \\ \midrule 
         MonoScene\(^\ast\) & 36.86 & 11.08 & 56.52 & 26.72 & 14.27 & 0.46  & 14.09 & 23.26 & 6.98  & 0.61  & 0.45  & 1.48  & 17.89 & 2.81  & 29.64 & 1.86  & 1.20  & 0.00  & 5.84  & 4.14  & 2.25 \\
        TPVFormer & 35.61 & 11.36 & 56.50 & 25.87 & \underline{20.60} & 0.85  & 13.88 & 23.81 & 8.08  & 0.36  & 0.05  & 4.35  & 16.92 & 2.26  & 30.38 & 0.51  & 0.89  & 0.00  & 5.94  & 3.14  & 1.52 \\
        VoxFormer\(^\dagger\) & 44.15 & 13.35 & 53.57 & 26.52 & 19.69 & 0.42 & 19.54 & 26.54 & 7.26 & 1.28 & 0.56 & 7.81 & \underline{26.10} & 6.10 & \underline{33.06} & 1.93 & 1.97 & 0.00 & 7.31 & 9.15 & 4.94 \\
        OccFormer & 36.50 & 13.46 & \underline{58.85} & 26.88 & 19.61 & 0.31  & 14.40 & 25.09 & \textbf{25.53} & 0.81  & 1.19  & 8.52  & 19.63 & 3.93  & 32.62 & 2.78  & \underline{2.82}  & 0.00  & 5.61  & 4.26  & 2.86 \\
        Symphonize & 41.92 & \underline{14.89} & 56.37 & 27.58 & 15.28 & 0.95  & 21.64 & \underline{28.68} & \underline{20.44} & \textbf{2.54} & \underline{2.82}  & \underline{13.89} & 25.72 & \underline{6.60} & 30.87 & \textbf{3.52}  & 2.24  & 0.00  & 8.40  & \textbf{9.57}  & 5.76 \\
        HASSC\(^\dagger\) & \underline{44.58} & 14.74 & 55.30 & \textbf{29.60} & \textbf{25.90} & \textbf{11.30} & \textbf{23.10} & 23.00 & 2.90 & 1.90 & 1.50 & 4.90 & 24.80 & \textbf{9.80} & 26.50 & 1.40 & \textbf{3.00} & 0.00 & \textbf{14.30} & 7.00 & \textbf{7.10} \\ 
         \midrule
         Ours & \textbf{44.98} & \textbf{15.73} & \textbf{61.58} & \underline{28.65} & 20.26 & \underline{1.03} & \underline{21.95} & \textbf{31.65} & 15.39 & \underline{2.36} & \textbf{3.42} & \textbf{14.58} & \textbf{26.73} & 6.52 & \textbf{35.20} & \underline{2.83} & 2.22 & 0.00 & \underline{9.26} & \underline{9.37} & \underline{5.80} \\ \bottomrule
    \end{tabular}
    \caption{Quantitative results on SemanticKITTI val set. \(^\ast\) represents the reproduced results in related papers \cite{tpvformer, occformer}. \(^\dagger\) denotes the results with temporal inputs. The best and the second-best results are in bold and underlined respectively.}
    \label{tab:semanticval}
\end{table*}

\begin{figure*}[!h]
    \centering
    \includegraphics[width=\linewidth]{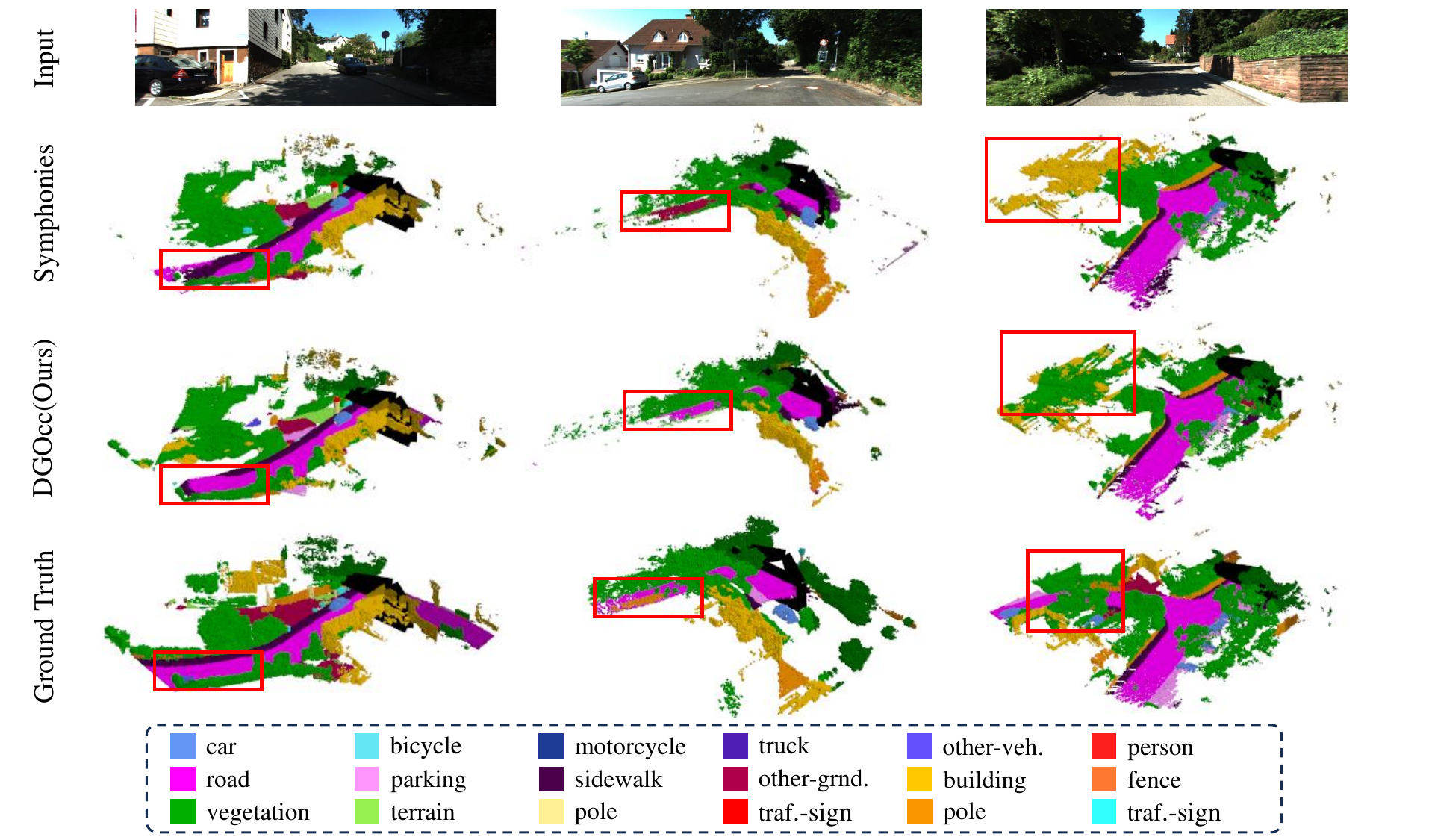}
    \caption{Qualitative results of Symphonies and DGOcc on SSCBench-KITTI-360 validation set. Our method possesses stronger hallucination capability and provides more accurate classification, thus constructing more complete 3D scenes.}
    \label{fig:visualization360}
\end{figure*}

\subsection{Implementation Details}
For the training of occupancy network on SemanticKITTI \cite{semantickitti} dataset, we only use the data of SemanticKITTI train set to pre-training. As for SSCBench-KITTI-360 \cite{kitti360, SSCBench}, because its LiDAR points don't have semantic annotation, we only use the LiDAR points to provide 2D depth labels. In order to provide semantic labels, we adopt the 2D semantic segmentation ground truth from KITTI-360 \cite{kitti360} dataset. Although the annotation formats of two datasets are different, the resulting pre-trained weights are still effective.

We pre-train the networks on 2 NVIDIA TITAN RTX GPUs for 30 epochs, with a batch of 8 samples. To align with our occupancy network, we crop the input images to \(1220\times 370\) and \(1396\times 372\) for SemanticKITTI and SSCBench-KITTI-360 datasets respectively. Random horizontal flip augmentation is implemented for both datasets. The AdamW \cite{adam} optimizer is adopted with a learning rate of 2e-4 for SemanticKITTI and 1e-4 for SSCBench-KITTI-360. The weight decay is 1e-4 for both datasets. Learning rate reduction occurs by a factor of 0.1 at the 25th epoch. We set \(dmax=192\) for both datasets.

\section{Additional Experimental Results}

\subsection{Quantitative Results on SemanticKITTI val.}
The results of quantitative experiments on SemanticKITTI \cite{semantickitti} val set are presented in Table~\ref{tab:semanticval}. The proposed method achieves the best performance with 44.98 IoU and 15.73 mIoU, exhibiting the effectiveness of the proposed method.

\subsection{Qualitative Results on SSCBench-KITTI-360 val.}
We provide more visual comparisons of DGOcc with Symphonies \cite{symphonies} on SSCBench-KITTI-360 \cite{kitti360,SSCBench} validation set, as depicted in Figure~\ref{fig:visualization360}. The red box in the first column shows that our method hallucinates more complete roads than Symphonies. The red boxes in the last two columns indicate that our approach achieves correct classification while Symphonies not. The extra visualization results confirm the efficacy of our approach once again.
\begin{table}[h]
    \centering
    \begin{tabular}{cc|cc|cc} \toprule 
        \multicolumn{2}{c|}{Input} & \multicolumn{2}{c|}{Supervision} & \multicolumn{2}{c}{Metric} \\ \midrule
        Depth & RGB & Dep. & Sem. & IoU\(\uparrow\) & mIoU\(\uparrow\) \\ \midrule
         & &  & & 44.55  & 15.65 \\
        \checkmark & \checkmark & & & 44.56 & 15.40 \\ 
        \checkmark &  & \checkmark & & \textbf{45.39} & 15.46 \\ 
         & \checkmark & \checkmark & \checkmark & 44.23 & 15.18 \\ \midrule
        \checkmark & \checkmark & \checkmark & \checkmark & 44.98 & \textbf{15.73} \\ \bottomrule
    \end{tabular}
    \caption{Ablation studies on the Depth Feature Extractor. Depth means the input prior depth maps while RGB indicates the input images. Dep. and Sem. represent depth supervision and semantic supervision respectively during the pre-training phase.}
    \label{tab:depthbranch}
\end{table}

\begin{table}[hb]
    \centering
    \begin{tabular}{cc|ccc} \toprule 
        \(\lambda_1\) & \( \lambda_2\) & IoU\(\uparrow\) & mIoU\(\uparrow\) & Rec.(\%) \\ \midrule 
        1.0 & 0.0 & 42.34 & 15.15 & 41.04\\
        1.0 & 0.1 & 44.14 & 15.01 & 41.35\\ 
        1.0 & 0.3 & \textbf{44.55} & \textbf{15.65} & 41.46\\ 
        1.0 & 0.5 & 44.17 & 15.15 & 41.09\\ 
        1.0 & 1.0 & 43.94 & 15.49 & 41.37\\ \bottomrule
    \end{tabular}
    \caption{Ablation studies on the low-level loss. The experiments are conducted without the Depth Feature Extractor. \(\lambda_1\) and \( \lambda_2\) are the weights of cross-entropy loss and multi-class version of Scene-Class Affinity Loss respectively. Rec. represents the Recall of voxels that require subdivision.}
    \label{tab:lambda}
\end{table}

\subsection{Additional Ablation Studies}

\noindent\textbf{Ablation on the Depth Feature Extractor: }
To validate the importance of Depth Feature Extractor, we present the ablation studies in Table~\ref{tab:depthbranch}. Adding Depth Feature Extractor without pre-training impairs mIoU slightly. It's because the Depth Feature Extractor is randomly initialized and the prior depth maps are noisy. With only depth pre-training, the IoU increases notably, implying that the explicit depth context information contributes to the geometry structure recovery. By adding additional semantic pre-training, the mIoU also improves. Semantic discriminability is augmented by introducing 2D semantic segmentation information. Additionally, when pre-training without estimated depth maps as inputs, IoU and mIoU both drop significantly, showing the importance of explicit depth context information contained in prior depth maps.

\noindent\textbf{Ablation on the low-level loss: }
To validate the impact of different components of the low-level loss, we implement ablation studies on corresponding loss weights, as presented in Table~\ref{tab:lambda}. The performance with only the cross-entropy loss is disappointing. After introducing the multi-class version of Scene-Class Affinity Loss, both IoU and mIoU increase notably, illustrating the pivotal importance and strong constraints of this loss. We choose \(\lambda_1=1.0\) and \(\lambda_2=0.3\) for the best balance between IoU and mIoU optimizations.

\end{document}